\documentclass[10pt,twocolumn,letterpaper]{article}

\usepackage{iccv}
\usepackage{times}
\usepackage{epsfig}
\usepackage{graphicx}
\usepackage{amsmath}
\usepackage{amssymb}

\usepackage{subcaption}
\usepackage{graphicx} % for pdf, bitmapped graphics files
\usepackage{multirow}
\usepackage{float}
\usepackage{authblk}
\usepackage{tikz}
\usepackage[pagebackref=true,breaklinks=true,letterpaper=true,colorlinks,bookmarks=false]{hyperref}

\iccvfinalcopy % *** Uncomment this line for the final submission

\def\iccvPaperID{6} % *** Enter the ICCV Paper ID here

\ificcvfinal\fi

\begin{document}
	
\newcommand\copyrighttextinitial{%
	
	\scriptsize This work has been submitted to the IEEE for possible publication. Copyright may be transferred without notice, after which this version may no longer be accessible.}%
\newcommand\copyrighttextfinal{%
	
	\scriptsize\copyright\ 2021 IEEE. Personal use of this material is permitted. Permission from IEEE must be obtained for all other uses, in any current or future media, including reprinting/republishing this material for advertising or promotional purposes, creating new collective works, for resale or redistribution to servers or lists, or reuse of any copyrighted component of this work in other works. DOI: 10.1109/ICCVW54120.2021.00111.}%
\newcommand\copyrightnotice{%
	
	\begin{tikzpicture}[remember picture,overlay]%
		
		\node[anchor=south,yshift=10pt] at (current page.south) {{\parbox{\dimexpr\textwidth-\fboxsep-\fboxrule\relax}{\copyrighttextfinal}}};%
	\end{tikzpicture}%
	
	%\vspace{-10pt}%
	
}

%%%%%%%%% TITLE
\title{Visual Domain Adaptation for Monocular Depth Estimation on Resource-Constrained Hardware}

\author[1]{Julia Hornauer}
\author[2]{Lazaros Nalpantidis}
\author[1]{Vasileios Belagiannis}
\affil[1]{Ulm University \\ Ulm, Germany \\ {\tt\small $\{$first.last$\}$@uni-ulm.de}}
\affil[2]{DTU -- Technical University of Denmark \\ Kgs. Lyngby, Denmark \\ {\tt\small lanalpa@elektro.dtu.dk}}

\maketitle
\copyrightnotice
\ificcvfinal\thispagestyle{plain}\fi

%%%%%%%%% ABSTRACT
\begin{abstract}
   Real-world perception systems in many cases build on hardware with limited resources to adhere to cost and power limitations of their carrying system. Deploying deep neural networks on resource-constrained hardware became possible with model compression techniques, as well as efficient and hardware-aware architecture design. However, model adaptation is additionally required due to the diverse operation environments. In this work, we address the problem of training deep neural networks on resource-constrained hardware in the context of visual domain adaptation. We select the task of monocular depth estimation where our goal is to transform a pre-trained model to the target's domain data. While the source domain includes labels, we assume an unlabelled target domain, as it happens in real-world applications. Then, we present an adversarial learning approach that is adapted for training on the device with limited resources. Since visual domain adaptation, i.e.~neural network training, has not been previously explored for resource-constrained hardware, we present the first feasibility study for image-based depth estimation. Our experiments show that visual domain adaptation is relevant only for efficient network architectures and training sets at the order of a few hundred samples. Models and code are publicly available\footnote{\url{https://github.com/jhornauer/embedded_domain_adaptation}}.
\end{abstract}

%%%%%%%%% BODY TEXT
\section{Introduction}

\begin{figure}[t]
	\centering
	\includegraphics[width=0.37\textwidth]{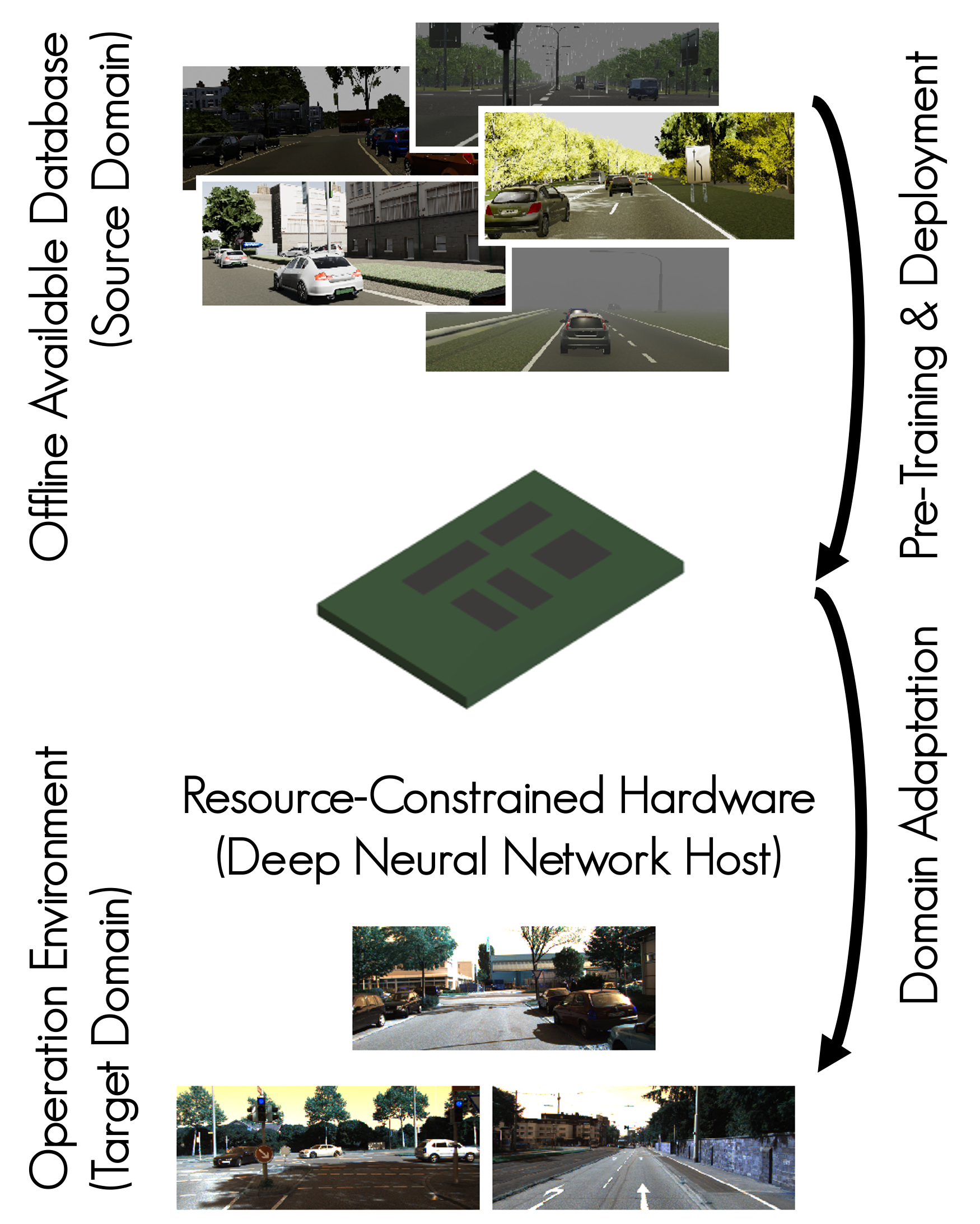}
	\caption{Resource-constrained hardware can be used for the deployment of deep neural networks. However, it is often required to further adapt the model because of the diverse operating environments. We study the problem of training a deep neural network on the embedded hardware in the context of domain adaption for monocular depth estimation. The visualized images are from the databases KITTI \cite{Geiger2013IJRR} and Virtual KITTI \cite{cabon2020vkitti2}.}
	\label{fig:teaser}
\end{figure}

Machine learning on resource-constrained hardware has emerged to an important research direction with applications on robotics~\cite{kim2019force}, autonomous driving~\cite{frickenstein2020binary}, and surveillance systems~\cite{sheng2013integrated}. Executing learning-based algorithms directly on-site reduces the system latency, preserves the data privacy, and makes the system more reliable because of the independence from external factors such as remote servers and communication networks. Moreover, resource-constrained hardware, such as embedded devices, is significantly less expensive than workstations or cloud services. Nevertheless, scaling up machine learning to real-world applications using resource-constrained hardware remains a challenge.

In practice, deep neural networks, the horsepower in the field, are very successful models for deployment on resource-constrained systems. To reach real-time performance, the model complexity and memory footprint are reduced with network compression~\cite{BelagiannisFG18}, as well as efficient~\cite{Howard2017MobileNetsEC} and hardware-aware~\cite{Zhang2020SkyNetAH} network architecture design. These approaches assume that further network training is not necessary after deployment. However, the integration of deep neural networks on low-cost embedded devices makes them more ubiquitous and at the same time exposes them to many and more diverse operation environments. Thus, on-device model adaptation is required for those devices to perform as expected. Training deep neural networks on the resource-constrained hardware, though, has not been addressed yet (illustrated in Fig.~\ref{fig:teaser}).

In this work, we address the problem of \textit{training deep neural networks on resource-constrained hardware} in the context of visual domain adaptation. Our testbed is monocular image-based depth estimation where the model adaptation from the source to the target domain happens in an unsupervised manner. We assume a pre-trained model that resulted from the source domain data and it has been trained with supervision. The model pre-training takes place on a standard workstation. For the target domain, we suppose that data collection is possible, e.g.~a mobile agent, but ground-truth depth maps are not available. Then, we present an adversarial learning approach~\cite{Kundu2018AdaDepthUC} that is adapted for training on the resource-constrained hardware. Given the hardware limitations, we employ an efficient network architecture~\cite{WofkFastDepthFM} for depth estimation. Besides, we also consider a complex architecture~\cite{LainaDeeperDepthP} for comparison reasons. Since domain adaptation has not been previously explored for resource-constrained devices, we present the first feasibility study for the perception task of monocular depth estimation. We analyze the training process of the deep neural network regarding the data and training set size, model complexity, and energy consumption. Our experiments show that visual domain adaptation on resource-constrained hardware---and thus deep neural network training---is meaningful only for efficient network architectures and training sets at the order of a few hundred samples.

\section{Related Works}
\textbf{Embedded depth estimation} deals with the depth prediction from a single image. In \cite{LainaDeeperDepthP}, \cite{Eigen2014DepthMP}, and \cite{Eigen2015PredictingDS}, promising results are shown based on deep neural networks for regression. However, none of these methods is designed for usage on resource-constrained devices. Instead, complex models have been proposed, which make hardware deployment challenging. In \cite{WofkFastDepthFM}, \cite{Poggi2018TowardsRU}, \cite{Oh2020RRNetRN} and \cite{8714893}, the effectiveness of different lightweight depth estimation network architectures is demonstrated on embedded devices, such as the Raspberry Pi and NVIDIA Jetson TX2. Poggi \textit{et al.}~\cite{Poggi2018TowardsRU} propose the lightweight architecture PyD-Net with a pyramidal feature extractor to train in an unsupervised manner for CPU processor usage. Wofk \textit{et al.}~\cite{WofkFastDepthFM} design their neural network, FastDepth, with depth-wise separable convolutions. Oh \textit{et al.}~\cite{Oh2020RRNetRN} propose a Repetition-Reduction block within the encoder, and a condensed decoding connection block for feature propagation to the decoder, in an encoder-decoder architecture. Peluso \textit{et al.} \cite{8714893} demonstrate their accuracy-driven quantization-aware training method adapted for the ARMv7 core on PyD-Net \cite{Poggi2018TowardsRU}. Nevertheless, their approaches address only the problem of deployment.

\textbf{Visual domain adaptation} refers to the generalization of a network trained on a source domain to some related target domain \cite{Csurka2017DomainAF}. In domain adaptation for depth-estimation, a major issue is the annotation of dense depth maps. In \cite{Abarghouei2018RealTimeMD}, \cite{Zheng2018T2NetST} and \cite{Zhao2019GeometryAwareSD} image-to-image translation is used to exploit data generation for addressing the problem. Instead of image-to-image translation Kundu \textit{et al.}~\cite{Kundu2018AdaDepthUC} explore an adversarial domain adaptation setting with two discriminators and different regularization techniques to obtain a monocular depth estimation model originally trained on synthetic data. Lasinger \textit{et al.}~\cite{Lasinger2020TowardsRM} target the generalization ability towards different domains by training their MiDaS network with multiple datasets of different scenes and environments. As most datasets differ in their depth ground truth representation, they create an objective that is unaffected by the various label types. Aleotti \textit{et al.} \cite{Aleotti2020RealtimeSI} create a large-scale dataset, called WILD, by making predictions on images of different environments with the pre-trained, large-scale MiDaS \cite{Lasinger2020TowardsRM} network. The resulting dataset is used to train lightweight models with a high generalization ability for deployment on handheld devices. We consider the adversarial domain adaptation~\cite{Kundu2018AdaDepthUC} as suitable for training on our hardware with limited resources. The aforementioned works accomplished great success in depth estimation and domain adaptation, but they rely on costly training of complex neural networks or only efficient model deployment on custom hardware.

\textbf{Resource-constrained hardware training} has not been considered at all. For example, lightweight architectures have been proposed in~\cite{WofkFastDepthFM}, \cite{Poggi2018TowardsRU} and \cite{Oh2020RRNetRN} with the aim to deploy models in real-time on embedded devices, where there is only a minor performance loss. More general, Zhang \textit{et al.}~\cite{Zhang2020SkyNetAH} propose to use hardware-aware neural network search to adapt the model for deployment to the dedicated hardware. Li~\textit{et al.}~\cite{Li2020BudgetTrain} propose linear learning rate scheduling with regard to limited training duration in terms of iteration. Although the design and deployment of hardware-aware and hardware-efficient networks has been studied in the past, the problem of training directly on the device with limited resources has not been addressed in those works.

\section{Approach}
We present the problem, our system and the domain adaptation algorithm for image-based depth estimation on resource-constrained hardware.

\subsection{Problem formulation}
Let $f_{\mathbf{\theta}}:\mathbb R^{w \times h \times 3} \rightarrow \mathbb R^{w \times h \times 1}$ to be the function that maps the image $\mathbf{x}$ with dimensions $w \times h \times 3$ to the depth map $\mathbf{y}$ with dimensions $w \times h \times 1$, where the function is represented by a deep neural network with parameters $\mathbf{\theta}$. The model $f_{\mathbf{\theta}}$ is trained with supervision on the database $\mathcal{S} = \{ (\mathbf{x},\mathbf{y})_{i} \}^{|\mathcal{S}|}_{i=1}$, which we refer to as the source domain. Consider now a different domain that is expressed by the data collection $\mathcal{T}=\{ (\mathbf{x})_{i} \}^{|\mathcal{T}|}_{i=1}$, referred to as the target domain. The target domain represents the operation environment. In the target domain, we assume not to have access to the ground-truth depth map $\mathbf{y}$. Furthermore, only a resource-constrained hardware system, e.g.~embedded device, is available for the model deployment. Then, our task is to adapt the model parameters $\mathbf{\theta}$ to the target domain without supervision, by relying only on the collected set $\mathcal{T}$ and the limited resources. Given the constrained hardware system and the visual domain adaptation task, we examine the training feasibility of the deep neural network w.r.t the data and training set size, as well as model complexity and the energy consumption. 

\subsection{Resource-constrained hardware}

We consider the NVIDIA Jetson Nano for our evaluations. The processing unit consists of the ARM-A57 CPU with 4GB RAM and the 128-core CUDA Maxwell GPU. The system is running a Linux-based operating system provided by NVIDIA and stored on a 128 GB SD card with up to 100 MB/s transfer speeds. The SD card memory is sufficient for the operating system, the executed libraries, developed algorithms and stored data. Finally, the domain adaptation is implemented in the PyTorch~\cite{paszke2019pytorch} framework. Note that we reckoned with the Raspberry Pi 4 for our experiments as well. However, the available processing power is not sufficient for training image-based deep neural networks in a reasonable time, since it is not equipped with a GPU.

\subsection{Visual domain adaption with limited resources}
We build the domain adaptation framework based on AdaDepth~\cite{Kundu2018AdaDepthUC}, an adversarial domain adaptation approach for depth estimation. This method is chosen because it relies on a less expensive setup with two discriminators instead of image-to-image translation as in the related approaches \cite{Abarghouei2018RealTimeMD}, \cite{Zheng2018T2NetST} and \cite{Zhao2019GeometryAwareSD}. We assume that the model $f_{\mathbf{\theta}}$ is composed of the encoder $\phi$ and the decoder $\psi$ network, such that:
\begin{equation}
	f_{\mathbf{\theta}} (\mathbf{x}) = \psi ( \phi ( \mathbf{x} )).
\end{equation}
The encoder $\phi$ maps the input image $\mathbf{x}$ to a latent space, whereas the decoder $\psi$ maps the latent space to the pixel-wise depth prediction.

At first, the encoder-decoder model $f_{\mathbf{\theta}}$ is pre-trained on the source domain database $\mathcal{S}$. Similarly to AdaDepth~\cite{Kundu2018AdaDepthUC}, $\psi$ is shared between the two domains, and thus it is not adapted. The domain adaptation transforms only the source domain encoder $\phi_{s}$ to the target domain encoder $\phi_{t}$. In practice, only a subset of the encoder parameters will be adapted, as we discuss later in Sec.~\ref{NetArch}. 

During training, we rely on the latent space discriminator and the depth map discriminator to distinguish between the latent space and the depth maps of the source and target domain respectively. The latent space discriminator $LD(\cdot)$ is trained to predict the domain of the latent space representations $\phi_{s}(\mathbf{x}_{s})$ and $\phi_{t}(\mathbf{x}_{t})$. Similar to LSGAN~\cite{mao2017least}, we define the objective as: 
\begin{equation}
	\begin{split}
		\mathcal{L}_{LD} = \mathbb{E}_{\mathbf{x}_{s} \sim \mathcal{S}}[\gamma (LD(\phi_{s}(\mathbf{x}_{s})) - 1)^{2}] +\\
		\mathbb{E}_{\mathbf{x}_{t} \sim \mathcal{T}}[\gamma (LD(\phi_{t}(\mathbf{x}_{t})))^{2}]+ \\
		\mathbb{E}_{\mathbf{x}_{t} \sim \mathcal{T}}[(1- \gamma) (LD(\phi_{t}(\mathbf{x}_{t})) - 1)^{2}],
	\end{split}
\end{equation}
where $LD(\phi_{t}(\mathbf{x}_{t})) - 1$ stands for the adversary, i.e.~setting the target domain as source domain (indicated by 1) and $\gamma \in \{0,1\}$ is used for updating the discriminators. The discriminator is updated when $\gamma=1$, while the encoder $\phi_{t}$ is updated for $\gamma = 0$ in a second step.

In addition, the depth map discriminator $DD(\cdot)$ takes the image and the depth map as input for identifying the domain. It is trained to distinguish the source ground-truth depth map $\mathbf{y}_{s}$ from the target's domain predicted depth map $\psi(\phi_{t}(x_{t}))$. The objective function for the depth map discriminator is given by: 
\begin{equation}
	\begin{split}
		\mathcal{L}_{DD} = \mathbb{E}_{(\mathbf{x}_{s}, \mathbf{y}_{s}) \sim \mathcal{S}}[ \gamma (DD(\mathbf{x}_{s} \mathbf{y}_{s}) - 1)^{2}]+ \\ 
		\mathbb{E}_{\mathbf{x}_{t} \sim \mathcal{T}}[ \gamma (DD(\mathbf{x}_{t}, \psi(\phi_{t}(\mathbf{x}_{t}))))^{2}] +\\
		\mathbb{E}_{\mathbf{x}_{t} \sim \mathcal{T}}[(1- \gamma) (DD(\mathbf{x}_{t}, \psi(\phi_{t}(\mathbf{x}_{t}))) - 1)^{2}].
	\end{split}
\end{equation}
Finally, we rely on the \textit{domain consistency regularization} loss from AdaDepth~\cite{Kundu2018AdaDepthUC} as a measure for the prevention of mode collapse. It minimizes the distance of the source and target latent space representation based on the target images. It is given by:
\begin{equation}
	\mathcal{L}_{reg} = \mathbb{E}_{\mathbf{x}_{t} \sim \mathcal{T}}[(1- \gamma)||\phi_{s}(\mathbf{x}_{t}) - \phi_{t}(\mathbf{x}_{t})||_{1}].
\end{equation}
The adversarial training starts from the pre-trained model $f_{\mathbf{\theta}}$ on the source domain data. As it progresses, $\phi_{t}$ is trained to produce samples that seem to be originating from the source domain. The minimization of all objectives is expressed as:
\begin{equation}
	\arg \min_{\gamma=1, LD, DD} \min_{\gamma=0, \phi_{t}} \mathcal{L}_{LD} + \mathcal{L}_{DD} + \lambda \mathcal{L}_{reg},
\end{equation}
where the hyper-parameter $\lambda$ controls the influence of the regularization term. For every iteration, the first minimization updates the parameters of the depth map discriminator $DD(\cdot)$ and latent space discriminator $LD(\cdot)$, while the second minimization updates the parameters of the target encoder $\phi_{t}$. The regularizer, finally, is applied only once during the second minimization where $\gamma=0$. The training process takes place on the resource-constrained hardware.  

\subsection{Neural network architectures for depth estimation}\label{NetArch}
Two main limitations of the resource-constrained hardware are the computing power and the available memory. The standard depth estimation network architectures are computationally complex and memory demanding~\cite{LainaDeeperDepthP}. To address this problem, there have been recently proposed lightweight architectures for depth estimation~\cite{WofkFastDepthFM}. For our experiments, we consider both a lightweight and a complex network architecture: FastDepth and ResNet-UpProj respectively.

\paragraph{Lightweight architecture} FastDepth~\cite{WofkFastDepthFM} is built with MobileNet~\cite{Howard2017MobileNetsEC} as the encoder and a lightweight decoder with depth-wise separable convolutions, followed by nearest-neighbor interpolation for up-sampling. The network counts 3.93M parameters in total~\cite{WofkFastDepthFM}. Additional skip connections between the encoder and the decoder are added for feature propagation to compensate for the small number of parameters. In line with the concept of \cite{Kundu2018AdaDepthUC}, only the last four layers of the encoder are trained during the domain adaptation. 

\paragraph{Complex architecture}
Laina \textit{et al.} design a encoder-decoder depth estimation network with ResNet-50 as the encoder and up-projection blocks within the decoder~\cite{LainaDeeperDepthP}. The architecture is implemented with five up-projections blocks, similar to FastDepth~\cite{WofkFastDepthFM}, and has 63.6M parameters in total. This is a significantly larger number of parameters compared to FastDepth. Similar to~\cite{Kundu2018AdaDepthUC}, the 5-th ResNet block is only adjusted during the adaptation.

At last, the depth map discriminator $DD(\cdot)$ follows the PatchGAN \cite{Isola2017ImagetoImageTW} network structure, while the latent space discriminator $LD(\cdot)$ is a convolutional discriminator which we later present based on the evaluation.

\section{Evaluation}
In this section, we present the findings of our approach on visual domain adaptation on the resource-constrained hardware for the demanding task of monocular depth estimation. In our evaluation, we consider the scenarios of indoor, as well as outdoor environments where we rely on four standard benchmarks for depth estimation. In each scenario, we study the factors of image resolution and training set size, model complexity and energy consumption during training on the device with the limited resources. We report the mean performance after five runs for each experiment. 

\subsection{Indoor and outdoor benchmarks}
In the indoor evaluation, the NYU Depth v2 dataset~\cite{Silberman:ECCV12} serves as the source domain database, while the target domain is represented by the DIML/CVL RGB-D data set~\cite{Kim2018DeepMD}. In the outdoor scenario, the synthetic Virtual KITTI (vKITTI)~\cite{cabon2020vkitti2} is the source domain and the target domain is the KITTI database~\cite{Geiger2013IJRR}.

\paragraph{From NYU Depth v2 to DIML/CVL RGB-D (indoors)} The NYU Depth v2 (source domain) is an indoor dataset taken at 464 different scenes by a depth-sensing camera. The scenes are split into 249 scenes for training and 215 for testing. The images have a resolution of 480x640. We rely on the training set for creating the pre-trained model. The  DIML/CVL RGB-D database (target domain) provides a large-scale indoor dataset with 220k training images taken by a Microsoft Kinect v2 at 283 different scenes of 18 different categories. In addition, there is a smaller set of 1500 training images and 500 samples for testing. The images and aligned depth maps have a resolution of 756x1344. The large-scale set utilization is not realistic because of the limited computational power and memory space, and thus we select the smaller set for the target domain.

\begin{figure*}[!ht]
	\centering
	\begin{subfigure}{0.18\textwidth}
		\includegraphics[width=\textwidth]{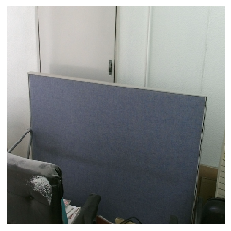}
		\includegraphics[width=\textwidth]{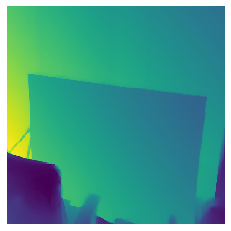}
		\caption{}
	\end{subfigure}
	~~
	\begin{subfigure}{0.18\textwidth}
		\includegraphics[width=\textwidth]{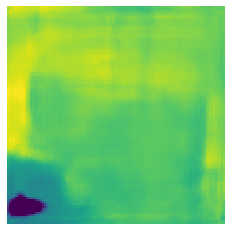}
		\includegraphics[width=\textwidth]{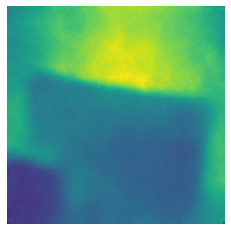}
		\caption{}
	\end{subfigure}
	\begin{subfigure}{0.18\textwidth}
		\includegraphics[width=\textwidth]{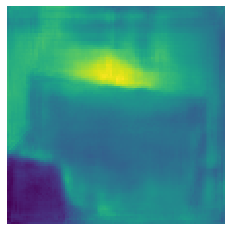}
		\includegraphics[width=\textwidth]{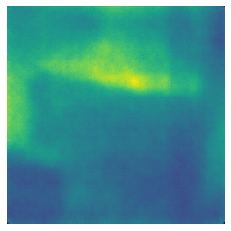}
		\caption{}
	\end{subfigure}
	\begin{subfigure}{0.18\textwidth}
		\includegraphics[width=\textwidth]{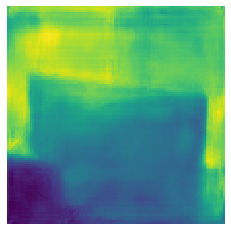}
		\includegraphics[width=\textwidth]{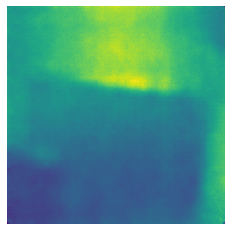}
		\caption{}
	\end{subfigure}
	\begin{subfigure}{0.18\textwidth}
		\includegraphics[width=\textwidth]{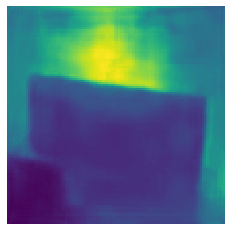}
		\includegraphics[width=\textwidth]{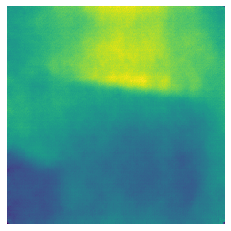}
		\caption{}
	\end{subfigure}
	\caption{Visual results on DIML/CVL RGB-D. (a): RGB (top) \& Ground truth (bottom). (b) - (e): Training FastDepth (top) \& ResNet (bottom) with a resolution of 224x224. (b) Source-only; (c) Domain adaptation for $|\mathcal{A}| = |\mathcal{B}| = 100$; (d) Domain adaptation for $|\mathcal{A}| = |\mathcal{B}| = 500$, (e) Domain adaptation for $|\mathcal{A}| = |\mathcal{B}| = 1000$.}
	\label{fig:visual_results_indoor}
\end{figure*}

\paragraph{From Virtual KITTI to KITTI (outdoors)} The synthetic virtual KITTI (vKITTI) dataset is used as the source domain for the outdoor scenario. It consists of 21260 synthetic image-depth pairs with a resolution of 375x1242, which we make use for training our networks. We rely only on the left images of the image-depth pairs. Since the maximum depth of KITTI is 80m, the ground truth depth is clipped to this value as maximum. The two examined architectures are pre-trained on vKITTI. On the other hand, the KITTI dataset is a real-world computer vision benchmark (target domain) with 42382 rectified stereo pairs. RGB image pairs with corresponding velodyne points are provided in the raw data, where we rely on the left images. The Eigen-split~\cite{Eigen2014DepthMP} is used for the evaluation since it is a common evaluation protocol in the literature \cite{Kundu2018AdaDepthUC}, \cite{Poggi2018TowardsRU}, \cite{Abarghouei2018RealTimeMD}, \cite{Zhao2019GeometryAwareSD}. Eigen \textit{et al.}~divide the data into 22600 samples for training, 888 samples for validation and 697 samples for testing.

\begin{table*}[!ht] 
	\caption{Evaluation of the indoor experiment training FastDepth and ResNet-UpProj with a resolution of 224x224. The results of the models trained on the source domain (source) and domain adaption with subsets $|\mathcal{A}| = |\mathcal{B}| \in \{100, 500, 1000\}$ are listed. Depth prediction results are evaluated on the DIML/CVL RGB-D test set. For accuracy $\delta$ higher is better, for RMSE lower is better. We also report the peak power consumption, average energy consumption per epoch, multiply–accumulate operations (MACs), as well as the average training duration per epoch (in minutes) and the inference time (in milliseconds) both for the Jetson device and for a PC workstation, for comparison.}
	\centering
	\begin{tabular}{ll|cccc|cc|c|cccc}
		\hline
		\multirow{2}{*}{\shortstack{Archi-\\tecture}} & Training & \multirow{2}{*}{$\delta_{1}$} & \multirow{2}{*}{$\delta_{2}$} & \multirow{2}{*}{$\delta_{3}$} & \multirow{2}{*}{RMSE} & Power & Energy & MACs & \multicolumn{2}{c}{Training [min]} & \multicolumn{2}{c}{Inference [ms]} \\
		& Data & & & & & [W] & [Wh] & [G] & Jetson & PC & Jetson & PC \\ 
		\hline
		\hline
		\multirow{4}{*}{\shortstack{Fast-\\Depth}} & Source & 0.493 & 0.847 & 0.958 & 0.824 & & & & & & \\
		\cline{2-13}
		% & $1500$ & 0.566 & 0.858 & 0.950 & 0.805 & 12.4 & 2.2 & \multirow{4}{*}{0.76} & 20.5 & 3.5 & \multirow{4}{*}{33} & \multirow{4}{*}{10} \\
		& $1000$ & 0.560 & 0.856 & 0.953 & 0.801 & 12.4 & 1.6 & \multirow{3}{*}{0.76} & 13.5 & 2.3 & \multirow{3}{*}{33} & \multirow{3}{*}{10} \\
		& $500$ & 0.563 & 0.861 & 0.954 & 0.796 & 11.8 & 0.8 & & 7 & 1.2 & & \\
		& $100$ & 0.562 & 0.862 & 0.955 & 0.803 & 11.3 & 0.2 & & 2.5 & 0.2 & & \\
		\hline
		\multirow{4}{*}{\shortstack{ResNet-\\UpProj}} & Source & 0.444 & 0.816 & 0.947 & 0.872 & & & & & & \\
		\cline{2-13}
		% & $1500$ & 0.581 & 0.858 & 0.943 & 0.782 & 12.3 & 14.0 & \multirow{4}{*}{32.25} & 91.5 & 7.1 & \multirow{4}{*}{610} & \multirow{4}{*}{44}  \\
		& $1000$ & 0.576 & 0.857 & 0.941 & 0.777 & 11.9 & 9.4 & \multirow{3}{*}{32.25} & 61.5 & 4.8 & \multirow{3}{*}{610} & \multirow{3}{*}{44} \\
		& $500$ & 0.578 & 0.858 & 0.942 & 0.755 & 11.9 & 4.6 & & 31 & 2.4 & & \\
		& $100$ & 0.573 & 0.861 & 0.951 & 0.749 & 11.9 & 1.0 & & 6.5 & 0.5 & & \\
		\hline
	\end{tabular}
	\label{tab:da_results_indoor}
\end{table*}

\begin{figure*}[ht]
	\centering
	\begin{subfigure}{0.18\textwidth}
		\includegraphics[width=\textwidth]{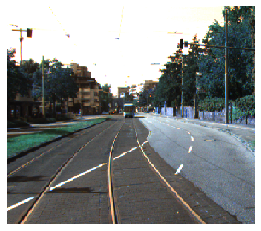}
		\vspace{0.1cm}
		\includegraphics[width=\textwidth]{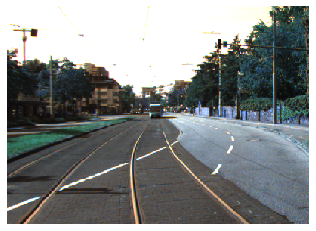}
		\caption{}
	\end{subfigure}
	~~
	\begin{subfigure}{0.18\textwidth}
		\includegraphics[width=\textwidth]{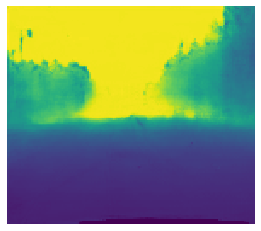}
		\vspace{0.1cm}
		\includegraphics[width=\textwidth]{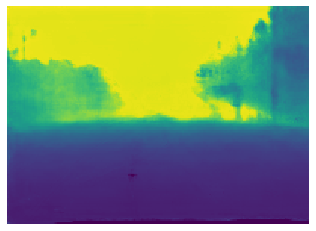}
		\caption{}
	\end{subfigure}
	\begin{subfigure}{0.18\textwidth}
		\includegraphics[width=\textwidth]{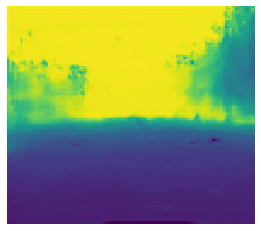}
		\vspace{0.1cm}
		\includegraphics[width=\textwidth]{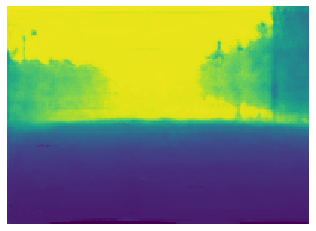}
		\caption{}
	\end{subfigure}
	\begin{subfigure}{0.18\textwidth}
		\includegraphics[width=\textwidth]{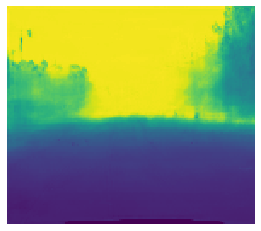}
		\vspace{0.1cm}
		\includegraphics[width=\textwidth]{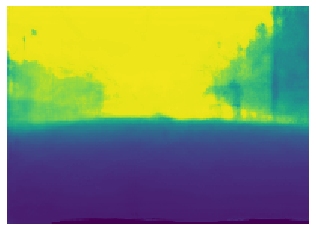}
		\caption{}
	\end{subfigure}
	\begin{subfigure}{0.18\textwidth}
		\includegraphics[width=\textwidth]{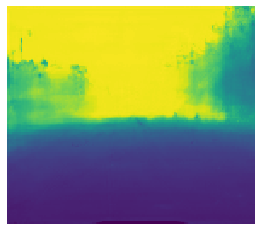}
		\vspace{0.1cm}
		\includegraphics[width=\textwidth]{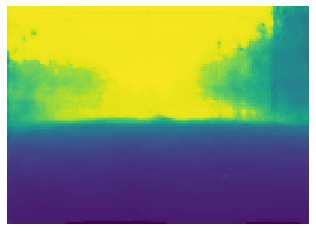}
		\caption{}
	\end{subfigure}
	\caption{Visual results on KITTI test image training FastDepth with a resolution of 256x512 (top) and 288x704 (bottom). (a) RGB; (b) Source-only; (c) Domain adaptation for $|\mathcal{A}| = |\mathcal{B}| = 100$; (d) Domain adaptation for $|\mathcal{A}| = |\mathcal{B}| = 500$; (e) Domain adaptation for $|\mathcal{A}| = |\mathcal{B}| = 1000$.}
	\label{fig:visual_results_outdoor}
\end{figure*}

\subsection{Training protocol \& implementation}
In both settings, the FastDepth~\cite{WofkFastDepthFM} and ResNet-UpProj~\cite{LainaDeeperDepthP} architectures are first trained on the source domains $\mathcal{S}$. The stochastic gradient descent (SGD) solver is used to optimize the $\mathcal{L}_{1}$ distance between the input image $\mathbf{x}$ and the depth map $\mathbf{y}$. Data augmentation, similar to \cite{WofkFastDepthFM}, is applied too. Random color jitter, random rotation, random scaling and center cropping is applied before the images are downsampled to a specific resolution. The final resolution for indoors is 224x224, while for outdoors we consider a lower 256x512 and a higher 288x704 resolution. The source domain training takes place on a PC workstation\footnote{The workstation has a 6-core processor with 16GB RAM and 6GB GPU memory.}. Then, the domain adaptation on both settings is performed using subsets of the training data $\mathcal{A} \subseteq \mathcal{S}$ and $\mathcal{B} \subseteq \mathcal{T}$ such that $|\mathcal{A}|=|\mathcal{B}|$. The subsets $\mathcal{A}$ and $\mathcal{B}$ of the training data $\mathcal{S}$ and $\mathcal{T}$ are selected randomly for every run. Different domain adaptation versions are trained using subsets of increasing size on the NVIDIA Jetson Nano. For all settings $\lambda$ is set to $0.7$. To imitate data collection on a real-world environment, we select small target domain sets such that $|\mathcal{A}| = |\mathcal{B}| \in \{100, 500, 1000\}$. Finally, the performance is evaluated on a target test set, which has not been observed during the domain adaptation.

\paragraph{Indoors}
The discriminator $LD(\cdot)$ consists of three convolutions with kernel size 3, each followed by a leaky rectified linear activation, and one last linear layer. After the last two convolutions dropout with probability 0.6 is applied. The two discriminators and the target encoder are optimized using the ADAM solver with $\beta_{1}$ = 0.5 and $\beta_{2}$ = 0.999. For FastDepth the learning rates are set to 0.0002 and for ResNet-UpProj to 0.00002. The augmentation is maintained during the adaptation in the indoor setting. Up to 25 epochs are necessary for the domain adaptation with FastDepth, i.e. 25 epochs for 100/500 samples and 15 epochs for the rest. With the ResNet-UpProj architecture it is 15 and 10 epochs respectively. One limiting factor of the embedded device is the available GPU memory. This leads to a maximum batch size depending on the model size. Training the adaptation with FastDepth a batch size of 16 per domain is selected. This is not possible with the larger ResNet-UpProj architecture, where the adaptation is trained with sub-batches of 2 samples per batch to maintain a resolution of 224x224.

\begin{table*}[t]
	\caption{Evaluation of the outdoor experiment training FastDepth with a resolution of 256x512 and 288x704. The results of the models trained on the source domain (source) and domain adaption with subsets $|\mathcal{A}| = |\mathcal{B}| \in \{100, 500, 1000\}$ are listed. The depth prediction results are evaluated on the KITTI test set. For accuracy $\delta$ higher is better, for RMSE lower is better. We also report the peak power consumption, average energy consumption per epoch, multiply–accumulate operations (MACs), as well as the average training duration per epoch (in minutes) and the inference time (in milliseconds) both for the Jetson device and for a PC workstation, for comparison.}
	\centering
	\begin{tabular}{ll|cccc|cc|c|cccc}
		\hline 
		\multirow{2}{*}{\shortstack{Reso-\\lution}} & Training & \multirow{2}{*}{$\delta_{1}$} & \multirow{2}{*}{$\delta_{2}$} & \multirow{2}{*}{$\delta_{3}$} & \multirow{2}{*}{RMSE} & Power & Energy & MACs & \multicolumn{2}{c}{Training [min]} & \multicolumn{2}{c}{Inference [ms]} \\
		& Data & & & & & [W] & [Wh] & [G] & Jetson & PC & Jetson & PC \\
		\hline
		\hline
		\multirow{4}{*}{256x512} & Source & 0.549 & 0.790 & 0.893 & 8.505 & & & & & & & \\
		\cline{2-13}
		& $1000$ & 0.649 & 0.825 & 0.906 & 9.337 & 11.8 & 3.2 & \multirow{3}{*}{1.98} & 23.4 & 2.3 & \multirow{3}{*}{37} & \multirow{3}{*}{10} \\
		& $500$ & 0.637 & 0.822 & 0.905 & 9.385 & 11.8 & 1.6 & & 11 & 1.3 & & \\
		& $100$ & 0.630 & 0.817 & 0.901 & 9.449 & 11.4 & 0.3 & & 2.4 & 0.3 & & \\
		\hline
		\multirow{4}{*}{288x704} & Source & 0.540 & 0.752 & 0.857 & 9.070 & & & & & & & \\
		\cline{2-13}
		& $1000$ & 0.652 & 0.825 & 0.906 & 9.567 & 11.8 & 5.2 & \multirow{3}{*}{3.07} & 34.5 & 3.1 & \multirow{3}{*}{38} & \multirow{3}{*}{10} \\
		& $500$ & 0.643 & 0.813 & 0.896 & 9.883 & 11.9 & 2.6 & & 17 & 1.6 & & \\
		& $100$ & 0.647 & 0.818 & 0.900 & 9.906 & 11.9 & 0.4 & & 3.5 & 0.3 & & \\
		\hline 
	\end{tabular}
	\label{tab:da_results_outdoor_fastdepth}
\end{table*}

\begin{table}[t]
	\caption{Evaluation of the outdoor experiment training FastDepth with a resolution of 256x512 and 288x704. For domain adaptation the results are reported with \textit{sample-wise median scaling} as in \cite{Kundu2018AdaDepthUC} and \cite{Zhou2017UnsupervisedLO}. The results of the models trained on the source domain (source) and domain adaption with subsets $|\mathcal{A}| = |\mathcal{B}| \in \{100, 500, 1000\}$ are listed. The depth prediction results are evaluated on the KITTI test set. For accuracy $\delta$ higher is better, for RMSE lower is better.}
	\centering
	\begin{tabular}{ll|cccc}
		\hline 
		\multirow{2}{*}{\shortstack{Reso-\\lution}} & Training & \multirow{2}{*}{$\delta_{1}$} & \multirow{2}{*}{$\delta_{2}$} & \multirow{2}{*}{$\delta_{3}$} & \multirow{2}{*}{RMSE} \\
		& Data & & & & \\
		\hline
		\hline
		\multirow{4}{*}{256x512} & Source & 0.549 & 0.790 & 0.893 & 8.505 \\
		\cline{2-6}
		& $1000$ & 0.649 & 0.863 & 0.938 & 7.813 \\
		& $500$ & 0.650 & 0.863 & 0.938 & 7.788 \\
		& $100$ & 0.637 & 0.854 & 0.930 & 7.942 \\
		\hline
		\multirow{4}{*}{288x704} & Source & 0.540 & 0.752 & 0.857 & 9.070 \\
		\cline{2-6}
		& $1000$ & 0.635 & 0.862 & 0.939 & 7.924 \\
		& $500$ & 0.626 & 0.856 & 0.934 & 7.952 \\
		& $100$ & 0.629 & 0.855 & 0.934 & 8.168 \\
		\hline 
	\end{tabular}
	\label{tab:da_results_outdoor_fastdepth_scaled}
\end{table}

\paragraph{Outdoors}
The discriminator $LD(\cdot)$ of the indoor adaptation is adopted, but the kernels in the convolutions are replaced by (4,7), (3,5) and (3,5) because of the differing resolution. The adaptation is performed for the resolutions 256x512 and 288x704. The outdoors resolution is too large for ResNet-UpProj, where we cannot run inference or training due to the limited resources. We rely only on FastDepth that is trained with the SGD optimizer with momentum set to 0.9 during the adaptation. The learning rates are set to \mbox{1e-4} or \mbox{1e-5} depending on the specific setup. The adaptation training requires up to 10 epochs, i.e. 10 epochs for 100/500 samples, and 5 epochs for the rest. The maximum possible batch size depends on the network parameters number. For both resolutions, FastDepth batch size  was 4. 

\subsection{Evaluation metrics and baselines}

The depth prediction is evaluated on the standard metrics of accuracy $\delta_{1}$, $\delta_{2}$ and $\delta_{3}$, as well as the root-mean-square error (RMSE), similar to the literature \cite{Zheng2018T2NetST}, \cite{Zhao2019GeometryAwareSD}. We report results for both network architectures by training only on the source domain (\textit{source}) and afterwards with domain adaptation. As an addition, for domain adaptation in the outdoor setting we follow another standard protocol in which the predicted depth maps are scaled by $median(y_{gt})/median(y_{pred})$ as in \cite{Kundu2018AdaDepthUC}, \cite{Zhou2017UnsupervisedLO}. The evaluation on the resource-constrained hardware is based on the peak power consumption, average energy consumption per epoch, multiply–accumulate operations (MACs), as well as the average training duration per epoch and the inference time both for the NVIDIA Jetson Nano (Jetson) device and for a PC workstation, as reference.

\subsection{Results and discussion}
We present the results for the indoor domain adaptation, followed by the outdoor configuration.

\paragraph{Indoors adaptation (NYU to DIML/CVL)}
In the indoor evaluation, four different subsets with $|\mathcal{A}| = |\mathcal{B}| \in \{100, 500, 1000\}$ of the training data are randomly selected to conduct the domain adaptation. Table \ref{tab:da_results_indoor} shows the performance results of FastDepth and ResNet-UpProj architectures on the DIML/CVL RGB-D test set and presents hardware-specific metrics and the model complexity. At first, it is clear that the domain adaptation is always helpful compared to only applying the model trained in the source domain (\textit{source}). Next, the lightweight FastDepth architecture functions with up to 1000 training data with faster training time per epoch than the ResNet-UpProj architecture. Moreover, the inference time of FastDepth is significantly faster. Also, the difference between the two architectures is large in the MAC operations. It is clear that the complex ResNet-UpProj architecture is more suitable for powerful computers. This also inferred when comparing with the workstation's training and inference time. Overall, relying on between $500$ to $1000$ training samples for adaptation results in a good balance between performance, training time and energy consumption. Finally, the complexity of the architecture does not play an important role in the energy consumption, which remains comparable for both models, considering that the more complex architecture converges faster. Fig.~\ref{fig:visual_results_indoor} illustrates the visual depth map results for one test sample. In this figure, for both architectures, the improvement of the model adaptation is visible (c) compared to training only on the source domain (b). Especially the lightweight architecture shows a significant improvement: the objects become distinguishable from the background and the object in the lower-left corner is predicted as closest.

\paragraph{Adaptation vKITTI to KITTI}
We follow the same configuration for this experiment as well. In Table~\ref{tab:da_results_outdoor_fastdepth}, we report the results of the domain adaptation in the resource-constrained hardware. In addition, we report the domain adaptation results with median-scaling in Table~\ref{tab:da_results_outdoor_fastdepth_scaled}. We rely now on two image resolutions, where the depth prediction results are in a similar range. The higher resolution adds considerable training time, but the inference time is similar. Furthermore, the energy consumption is not significantly affected by the input resolution. On the other hand, the complex ResNet-UpProj architecture is not capable of running on the NVIDIA Jetson Nano due to the memory limitation with either image resolutions. Finally, the results after adaptation overall improve for both protocols. Moreover, our visual results show improvement from source only to domain adaptation, as shown in Fig.~\ref{fig:visual_results_outdoor}.

\paragraph{Discussion}
Both indoor and outdoor evaluations demonstrate the feasibility of domain adaptation on the resource-constrained hardware in a meaningful period of time. The image resolution and number of samples of course affects the training time, which can go up to 61.5 minutes for ResNet (indoors) and 34.5 minutes for FastDepth (outdoors). The energy consumption is not a concern for any of our experiments, while the inference time is architecture dependent. For instance, we reach 10 milliseconds with FastDepth for both experiments. Thus, we conclude that training directly on the embedded device is possible with adversarial training. Given that monocular depth estimation is a demanding task, we expect perception tasks such as human trajectory estimation~\cite{hasan2019forecasting} and gesture recognition~\cite{wiederer2020traffic} or other robotics applications~\cite{nissler2015omg, engel2019deeplocalization} to be easier transferable to the resource-constrained hardware. Developing the complete perception of an autonomous agent on resource-constrained hardware is part of our future work. The main benefit will be to reduce the overall energy demands, while maintaining reliable performance.

\section{Conclusion}
We presented the first feasibility study on training deep neural networks on resource-constrained hardware in the context of visual domain adaptation. Our testbed is monocular depth estimation, where domain adaptation is accomplished without supervision. We extended an adversarial learning approach to function on the device with limited resources. In two evaluations using four standard databases, we have shown that domain adaptation on the resource-constrained hardware is manageable for lightweight architectures based on a few hundred samples from the target domain. Our study indicates that the deployment-hardware needs always to be considered along with the training algorithm, neural network architecture and the type of supervision to scale up machine learning to real-world applications.

{\small
\bibliographystyle{ieee_fullname}
\bibliography{mybibfile}
}

\end{document}